\documentclass{article} 
\usepackage{gdd,times}
\title{Diffusion Models Are Innate One-Step Generators}
\usepackage{graphicx,setspace,booktabs,amsfonts,nicefrac,microtype,tikz,mathtools, url, pbox, tabu, amsmath,algorithm,amsmath, amssymb, caption, subcaption, multirow, overpic, textpos, pifont, hyperref}

\newcommand{\NA}{---}
\newlength\savewidth\newcommand\shline{\noalign{\global\savewidth\arrayrulewidth
  \global\arrayrulewidth 1pt}\hline\noalign{\global\arrayrulewidth\savewidth}}
\newcommand{\tablestyle}[2]{\setlength{\tabcolsep}{#1}\renewcommand{\arraystretch}{#2}\centering\footnotesize}

\newcolumntype{x}[1]{>{\centering\arraybackslash}p{#1pt}}
\newcolumntype{y}[1]{>{\raggedright\arraybackslash}p{#1pt}}
\newcolumntype{z}[1]{>{\raggedleft\arraybackslash}p{#1pt}}

\newcommand{\app}{\raise.17ex\hbox{$\scriptstyle\sim$}}

\definecolor{deemph}{gray}{0.6}

\author{Bowen Zheng \\
  Institute of Neuroscience
  \\Key Laboratory of Brain Cognition and Brain-inspired Intelligence Technology
  \\Center for Excellence in Brain Science and Intelligence Technology
  \\Chinese Academy of Sciences\\
  Shanghai, China \\
\texttt{zhengbw@ion.ac.cn} \\
\And
Tianming Yang \thanks{Corresponding author.} \\
  Institute of Neuroscience
  \\Key Laboratory of Brain Cognition and Brain-inspired Intelligence Technology
  \\Center for Excellence in Brain Science and Intelligence Technology
  \\Chinese Academy of Sciences\\
  Shanghai, China \\
\texttt{tyang@ion.ac.cn}
}

\iclrfinalcopy 
\begin{document}
\maketitle
\begin{abstract}
Diffusion Models (DMs) have achieved great success in image generation and other fields. By fine sampling through the trajectory defined by the SDE/ODE solver based on a well-trained score model, DMs can generate remarkable high-quality results. However, this precise sampling often requires multiple steps and is computationally demanding. To address this problem, instance-based distillation methods have been proposed to distill a one-step generator from a DM by having a simpler student model mimic a more complex teacher model. Yet, our research reveals an inherent limitations in these methods: the teacher model, with more steps and more parameters, occupies different local minima compared to the student model, leading to suboptimal performance when the student model attempts to replicate the teacher. To avoid this problem, we introduce a novel distributional distillation method, which uses an exclusive distributional loss. This method exceeds state-of-the-art (SOTA) results while requiring significantly fewer training images. Additionally, we show that DMs' layers are differentially activated at different time steps, leading to an inherent capability to generate images in a single step. Freezing most of the convolutional layers in a DM during distributional distillation enables this innate capability and leads to further performance improvements. Our method achieves the SOTA results on CIFAR-10 (FID 1.54), AFHQv2 64x64 (FID 1.23), FFHQ 64x64 (FID 0.85) and ImageNet 64x64 (FID 1.16) with great efficiency. Most of those results are obtained with only 5 million training images within 6 hours on 8 A100 GPUs.

\end{abstract}

\section{Introduction}

\begin{figure}[htbp]
\centering
    \includegraphics[width=\linewidth]{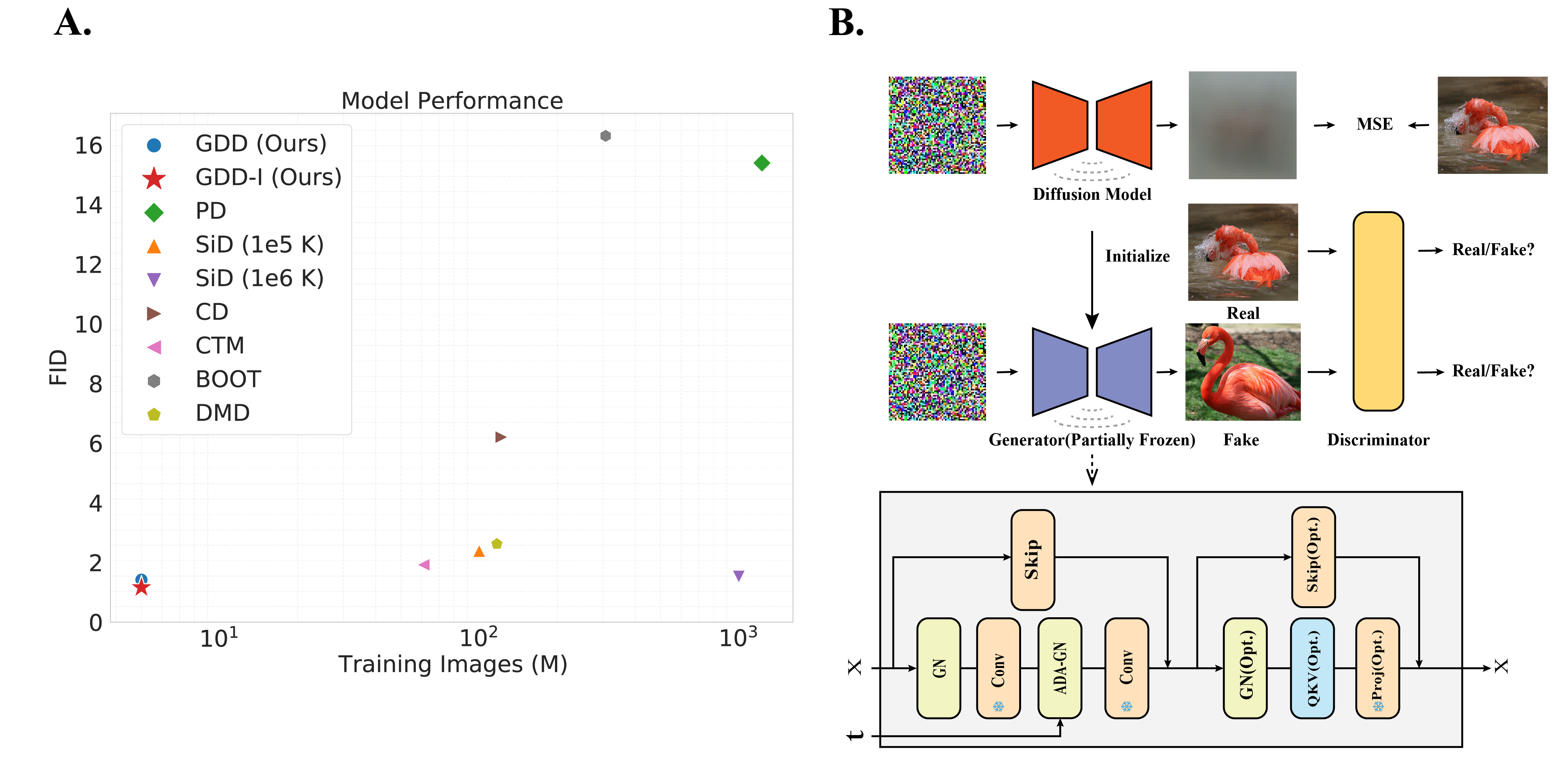}
   \caption{Proposed method and comparison of other methods on ImageNet 64x64. \textbf{A.} Comparison on ImageNet 64x64. Our methods achieve SOTA results with fewer training data. \textbf{B.} Proposed method. We initialize generator with a pre-trained diffusion model and freeze most of the convolutional layers, then adopt exclusively a distributional level loss. }
\label{fig:MAIN}
\end{figure}

Diffusion Models (DMs, \cite{song2020generative},\cite{song2020improved},\cite{ho2020denoising}) have led to an explosion of generative vision models. By sampling through the trajectory defined by Probability Flow Ordinary Differential Equation (PF ODE) or Stochastic Differential Equation (SDE) solvers (\cite{song2021scorebased}), DMs can generate high-quality images and videos, surpassing many traditional methods, including Generative Adversarial Networks (GAN, \cite{goodfellow2014generative}). However, to produce high-quality results, DMs require multiple sample steps to generate an accurate trajectory, which is computationally expensive. To alleviate this problem, a series of distillation methods (\cite{salimans2022progressive}, \cite{song2023consistency}, \cite{kim2024consistency}) have been proposed to distill a one-step generative model. However, these methods often require substantial computational resources, yet produce poorer performance than original models.

In this work, we first demonstrate that the underlying limitation of these instance-based distillation methods is due to the fact that the local minima for the teacher model and the student model can be very different. We then introduce a distributional distillation method based exclusively on GAN loss. Our method is highly efficient, reaching state-of-the-art (SOTA) results with only 5 million training images. In comparison, instance-based distillation methods often require hundreds of millions of training images. 

Furthermore, we explore the mechanism underlying the efficient distillation of our method. Our analyses indicate that the layers in DMs are differentially activated at different time steps, suggesting that DMs inherently possess the capability to generate images in a single step. Freezing most of the layers in a diffusion model leads to even better distillation results.

In summary, our approach achieves SOTA results by a significant margin with fewer training images (Figure \ref{fig:MAIN}). Our findings not only introduce a novel approach that achieves SOTA results with drastically reduced computational cost, but also offer valuable insights for further investigations of diffusion distillation.
\label{sec:intro}

\section{Limitation of Instance-based Distillation}\label{sec:instance_intro}
\subsection{Instance-based Distillation}
Given a pre-trained score model \(\mathbf{g}_{\phi}(x_{t_{i}}, t_i)\), where \(x_{t_{i}} \sim \mathcal{N}(x_0, t_{i}^2 I) \) is a perturbed data point, \(x_0\) is the clean image, \(t_i\) is the standard deviation of the perturbed distribution with a Gaussian kernel and sampled from a time-step scheduler function \(T(i,N)\), \(N\) is total discrete steps, and \(i\) is index of current sample step, we can define an ODE solver \(\mathbf{Solver_{\phi}}\):
\begin{equation} 
\displaystyle \mathbf{Solver}_{\phi}(x_{t_{i}},t_{i},t_{i-1}) = \frac{t_{i}-t_{i-1}}{t_{i}}(\mathbf{g}_{\phi}(x_{t_{i}},t_{i}) - x_{t_{i}})+x_{t_{i}}
\end{equation}
There are many different types of \(\mathbf{Solver}_{\phi}\), e.g., Heun solver introduced in EDM (\cite{karras2022elucidating}). We use Euler solver as an example here. Then we can sample iteratively with this solver to get the final results.

An instance-based distillation typically has a series of teacher models \(\mathbf{H}\) and student models \(\mathbf{F}_\theta\). These teacher models often have more steps than student models. For example, in Progressive Distillation (PD, \cite{salimans2022progressive}), teacher models are defined as 
\begin{equation}
    \mathbf{H} =\begin{cases}
        \mathbf{Sovler}_{\phi}(\mathbf{Solver}_{\phi}((x_{t_i},t_i,t_j), t_j, t_k), & i-j=1 , j-k=1 \\
        \mathbf{F}^{sg}_{\theta}(\mathbf{F}^{sg}_{\theta}((x_{t_{i}},t_{i},t_{j}), t_{j},t_k ), & i-j\neq 1 , j-k\neq 1
    \end{cases}
\end{equation}
where \(sg\) is the stop gradient to prevent gradient leak to the teacher models. More detailed introduction to other instance-based distillation methods can be found in Appendix \ref{sec:instance_details}. 

In these instance-based distillation methods, a teacher model \(\mathbf{H}\) contains more parameters than the student and has to forward through the neural network multiple times, while the student model \(\mathbf{F}_\theta\) has less parameters and passes through the neural network in a single iteration.

\subsection{Inconsistent local minimum}\label{sec:instance}

\begin{figure}[hb]
\centering
    \includegraphics[width=\linewidth]{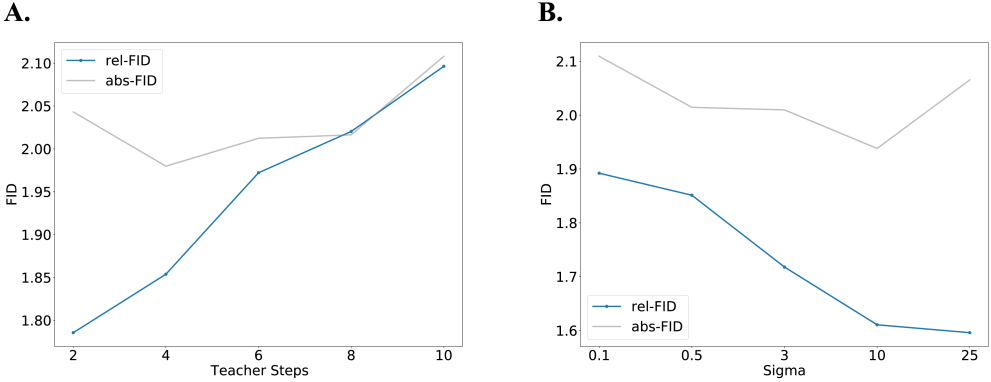}
   \caption{rel-FIDs show strong correlation with the steps and sigmas. \textbf{A.} rel-FIDs (blue) and abs-FIDs (grey) for teacher models with different steps. \textbf{B.} The abs-FIDs (grey) and rel-FIDs (blue) for two-step teacher models with different sigma of intermediate point. FIDs are computed from 50,000 images generated by each model with fixed seeds from 0 to 49,999, and an inception model is used to generate features required for calculating FID. }
\label{fig:diff}
\end{figure}

The difference between the teacher and the student leads to a significant inductive bias: the teacher model can transform between pixel space and latent space several times while the student model can only perform the transformation once. We speculate this may produce different optimization landscapes and different local minima between the teacher and student model. Forcing the student model to approximate the teacher model's results may, therefore, yield sub-optimal results.

To verify our hypothesis, we use the Fréchet Inception Distance (FID, \cite{heusel2018gans}) to compare between two image sets. We term the FID computed between the teacher and the student model as relative FID (rel-FID), and the FID computed against the training-set images as absolute FID (abs-FID). 

We define a series of teacher models with 2, 4, 8, 10 steps with PD manner. For example, two-step teacher model is parameterized as \(H=F_\theta(F_\theta(x_{t_2},t_2,t_1),t_1,t_0))\) where \(t_i \in \{T(2,2), T(1,2), T(0,2)\}\) , and so on. We train the student model and those teacher models with GAN loss alone. Several training techniques are used (See Appendix \ref{sec:rel-FID_details} for more details). Finally, we compute the FID (rel-FID) between the student model and each teacher model following the evaluation protocol in EDM (\cite{karras2022elucidating}). Figure \ref{fig:diff} \textbf{A.} shows the abs-FIDs and rel-FIDs for the teacher models with different numbers of steps. While abs-FID of those teacher models are similar, the teacher models with more steps have larger rel-FID, suggesting a larger difference compared to the student model. When we fix the teacher models' steps to two and adjust the sigma of the intermediate time step, the rel-FID decreases (Figure \ref{fig:diff} \textbf{B.}). Teacher models with higher sigmas produce images more similar to those generated by the one-step student model.

These results reveal that the teacher and student models may achieve similar performance, measured by abs-FID, but in different ways. Teachers with fewer steps are more similar to the one-step student, but there are still a significant difference between them (rel-FID=1.78 for a two-step teacher). The student model fails to mimic the teacher model at the instance level which might be the reason that limits the performance of instance-based distillation methods.

\section{Distilling DMs with Distributional loss }
To circumvent the limitations in the instance-based distillation methods, we can supervise the student model with a distribution-level loss instead of asking the student model to mimic the teacher at the instance level.

GAN loss is one of the most commonly used distribution-level loss. Distilling DMs with GAN loss have been introduced in previous work for different purposes. GAN loss is applied as an auxiliary loss after the instance-based distillation to improve its performance in Consistency Trajectory Model (CTM, \cite{kim2024consistency}) as originally introduced in Vector-Quantized GAN (VQ-GAN, \cite{esser2021taming}). SDXL-Lightning (\cite{lin2024sdxllightning}) uses GAN loss to replace the deterministic instance-level loss (L2 or LPIPS) with a probability form, but the discriminator still receives instance-level information about the time step and the beginning point of the trajectory to ensure that the student follows the original trajectory. Distribution Matching Distillation (DMD, \cite{yin2023onestep}) adopts another type of distributional distillation method, which uses a distribution matching gradient instead of GAN loss. But again, an instance-level pair-matching loss is added alongside the GAN loss. Although these works do not use distributional loss alone, they demonstrate the potential of distributional-loss based distillation.

\subsection{GAN Distillation at Distribution level}
For distilling a one-step generator from DMs, we introduce a new method called GDD (GAN Distillation at Distribution Level) , which uses only a distribution-level loss WITHOUT any instance-level supervision. Importantly, unlike in SDXL-Lightning or DMD, our method uses real data instead of the data generated by the teacher model as the ground truth. We use non-saturating GAN objective defined as:
\begin{equation}\label{eq:GANobjective}
\begin{aligned}
      &\max_{\mathbf{D}}~\mathbb{E}_{\mathbf{x}} [\log (\mathbf{D}(\mathbf{x}))] + \mathbb{E}_{\mathbf{z}}[1-\log(\mathbf{D}(\mathbf{G}_{\theta}(\mathbf{z})))]\\
     &\min_{\mathbf{G_{\theta}}}\mathbb{E}_{\mathbf{z}}[-\log(\mathbf{D}(\mathbf{G}_{\theta}(\mathbf{z})))]
\end{aligned}
\end{equation}

With a pre-trained diffusion model, the score model \(\mathbf{g_\phi}\) is used as the generator \(\mathbf{G}_{\theta}\) and a pre-trained VGG16 (\cite{simonyan2015deep}) is used as a discriminator by default.

\subsection{Experiments}\label{sec:EXP1}

\begin{table*}[t]
\vspace{-.2em}
\centering
\subfloat[
\textbf{Discriminators}. Discriminators that are either too strict or too weak can lead to the collapse of the model. Multi-scale discriminators (LPIPS and PG) show better performance and the discriminator proposed in Projected GAN achieves the best FID. 
\label{tab:discriminator}
]{
\begin{minipage}{0.35\linewidth}
\begin{center}
\tablestyle{6pt}{1.05}
\begin{tabular}{y{90}x{28}}

Discriminator & FID (\(\downarrow\)) \\
\shline
StyleGAN2 (Scratch) & collapse \\
StyleGAN2 (Pre-trained) & collapse \\
VGG16 & 4.04 \\
LPIPS &  3.49  \\
PG &  \textbf{2.21}   \\

\end{tabular}
\end{center}
\end{minipage}}
\hspace{4em}
\subfloat[
\textbf{Augmentation and Regularization}. Augmentation leads to poor results, but r1 regularization improves performance significantly.
\label{tab:aug_reg}
]{
\centering
\begin{minipage}{0.32\linewidth}
\begin{center}
\tablestyle{4pt}{1.05}
\begin{tabular}{x{54}x{24}x{28}}
Augmentation & \(\gamma_{r1}\) & FID (\(\downarrow\)) \\
\shline
\ding{51} & \NA & 2.21 \\
\ding{55} & \NA & 1.98 \\
\ding{55} & 1e-3 & 1.77 \\
\ding{55} & 1e-4 & \textbf{1.66} \\
\ding{55} & 1e-5 & 1.66 \\
\end{tabular}
\end{center}
\end{minipage}}
\hspace{2em}

\vspace{-.1em}
\caption{\textbf{GDD with different discriminators, data augmentations, and regularizations} on CIFAR-10. EDM with NCSN++ is used as the generator. Results are based on 5 million training images.}
\label{tab:ablations} \vspace{-.5em}
\end{table*}

\paragraph*{Baseline.}
We perform our experiments on CIFAR-10 (\cite{krizhevsky2009learning}) as the baseline. EDM with NCSN++ (\cite{song2021scorebased}) architecture is used as the generator. We choose VGG16 (\cite{simonyan2015deep}) as the baseline discriminator. Adaptive augmentation (ADA, \cite{karras2020training}) is adopted with r1 regularization (\cite{mescheder2018training}) where \(\gamma_{r1}=0.01\) as suggested in StyleGAN2-ADA (\cite{karras2020training}). Our baseline setting shows good performance on CIFAR-10 dataset with only 5M training images and already achieves performance (FID=4.04) close to that of the Consistency Distillation (FID=3.55, \cite{song2023consistency}).

\paragraph{Discriminator.}
An appropriate discriminator is crucial since it represents the overall target distribution and determines the performance of the generator.  We compare the StyleGAN2 (Both scratch and pre-trained with GAN objective, \cite{karras2020analyzing}), VGG, LPIPS (\cite{zhang2018unreasonable}) and Projected GAN (PG, \cite{sauer2022styleganxl}) discriminators. Adaptive augmentation (ADA) is adopted on StyleGAN2, VGG and LPIPS discriminators with r1 regularization where \(\gamma_{r1}=0.01\). 

For the PG Discriminator, we use a non-saturating version of PG objective: 
\begin{equation}
\begin{aligned}
    &\max_{\{\mathbf{D}_l\}} \sum_{l \in \mathcal{L}} \Big ( 
    \mathbb{E}_{\mathbf{x}} [\log \mathbf{D}_l(\mathbf{P}_l(\mathbf{x}))] + \mathbb{E}_{\mathbf{z}}[1-\log(\mathbf{D}_l(\mathbf{P}_l(\mathbf{G}_{\theta}(\mathbf{z}))))] \Big)\\
    &\min_\mathbf{G_{\theta}}\sum_{l \in \mathcal{L}} \Big ( \mathbb{E}_{\mathbf{z}}[ -\log(\mathbf{D}_l(\mathbf{P}_l(\mathbf{G}_{\theta}(\mathbf{z})))) \Big)
\end{aligned}
\end{equation}
\label{eq:GANobjective2}

We use VGG16 with batch normalization (VGG16-BN, \cite{simonyan2015deep}, \cite{ioffe2015batch}) and EfficientNet-lite0 (\cite{tan2020efficientnet}) as feature networks \(\mathbf{P}\). We adopt differentiable augmentation (diffAUG, \cite{zhao2020differentiable}) without gradient penalty by default, following original work.

As illustrated in Table \ref{tab:discriminator}, discriminators with multiple scales are better than the vanilla VGG discriminator. Overall, the PG discriminator with a fusion feature based on VGG16-BN and EfficientNet-lite0 achieves the best results. Therefore, we will use the PG discriminator and objective in the following experiments.

\paragraph*{Augmentation and Regularization.}
Differentiable augmentation (diffAUG) was introduced in \cite{zhao2020differentiable} to prevent the overfitting of discriminators, but we find that it leads to poor results in our method (Table \ref{tab:aug_reg}). This may be because EDMs are pre-trained with data augmentation. We disable all augmentation in all of our further experiments. In the original PG, regularization for the discriminator was not used. However, our experiments show that it is important to apply r1 regularization on the discriminator to prevent overfitting and it improves performance significantly (Table \ref{tab:aug_reg}). The value of \(\gamma_{r1}\) should be kept small since the discriminator of the Projected GAN generates multiple output logits, and a large \(\gamma_{r1}\) may lead to a numerical explosion. With optimal settings, GDD reaches the SOTA results (FID=1.66) with only 5 million training images.

\section{The Innate Ability of DMs for One-Step Generation}\label{sec:Innate}
We have demonstrated that GDD is both efficient and produces high-quality results, but we wonder why it can learn to generate images in one step so rapidly. Training DMs with instance-based distillation often requires hundreds of millions of images, while a few million images are enough for GDD. 

We hypothesize that DM layers may handle the tasks in the diffusion objective relatively independently at different time steps. Finding a way to activate them together may let a DM achieve what normally requires many steps in one step. If so, we should observe that layers in diffusion models are differentially activated across time steps. Moreover, we can freeze most of the layers in DMs but train the model to learn how to use their innate capabilities to produce good results.

\subsection{Differential Activations across time steps}\label{sec:special}
\begin{figure}[htb]
\centering
   \includegraphics[width=\linewidth]{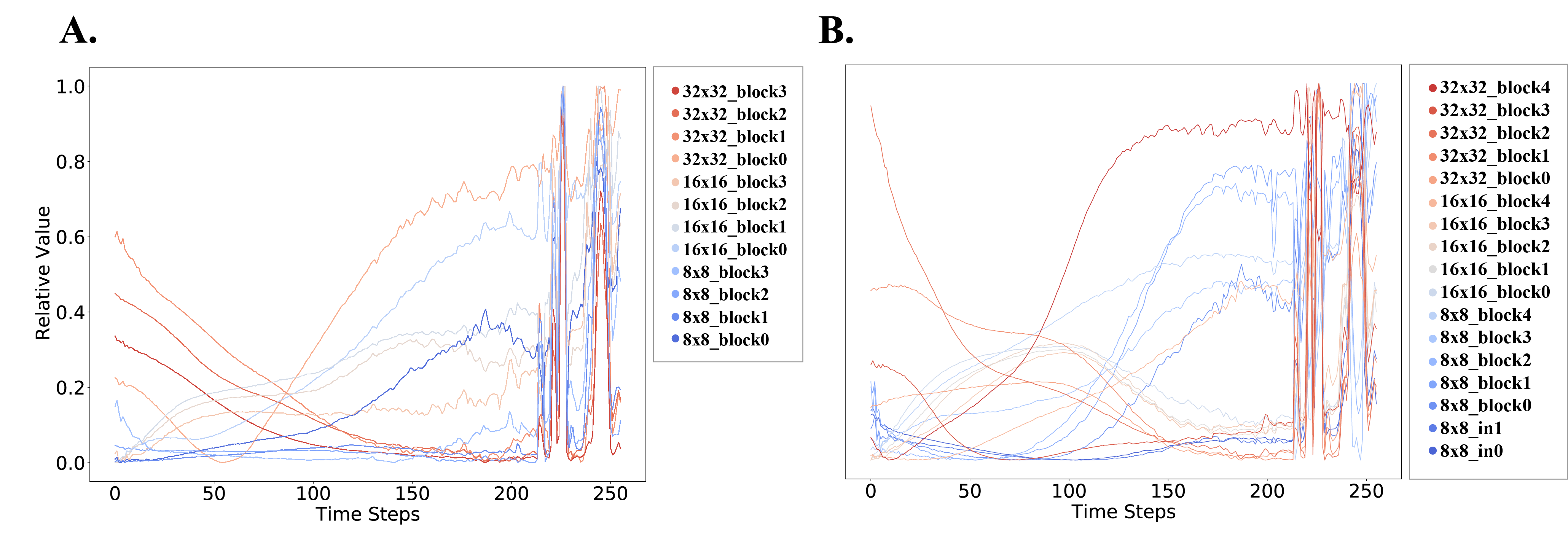}
\caption{The relative value of the last convolutional layer in each UNet block at different depth. Values are normalized with min-max normalization across time steps for each layer. Red-to-blue color gradient indicates shallow to deeper layers. \textbf{A.} Encoder. \textbf{B.} Decoder.}
\label{fig:EncDec}
\end{figure}
To analyze the activation of the inner layers, we record the outputs of each UNet block's last convolutional layer during sampling. To obtain more accurate and comprehensive data, we set the total number of time steps to 256 (\(N=256\)), where \(i=256\) is the beginning step with \(T(256,N)=\sigma_{max}\) and \(i=0\) is the end step with \(T(0,N)=\sigma_{min}\). Outputs are grouped by time steps and depths first, then min-max normalization are applied across time steps as 
\(\Hat{O}_{t}^{l} = \frac{O_{t}^{l} - \min(O^l)}{\max(O^l) - \min(O^l)}\). 
We find that the convolutional layers are differentially activated at different time steps except for the two periods in the beginning time steps at around \(\sigma=40\) (time step=225) and \(\sigma=60\) (time step=243), where almost all layers are activated. Other than these two periods, the shallow layers are maximally activated at the end time steps, while the deep layers are more sensitive to the beginning time steps. The middle layers are in between (Fig \ref{fig:EncDec}). These trends can be seen in both the encoder and the decoder, except that several shallow layers are sensitive to the beginning time steps. 

\subsection{Achieving One-Step Generation via Freezing}
The DM layers are differentially activated at different time steps. If we can find a way to activate them simultaneously for one-step generations, we can utilize what has already learned in these layers, their innate capabilities, and distill a DM quickly. As the group normalization layers (both the adaptive and the vanilla group normalizations) serve as the triggers of the convolutional layers, we may only need to tune them during the distillation. To test this idea, we examine the model's performance on CIFAR-10 with most of the convolutional layers frozen, while allowing the group normalization layers, the input layer and the output layer (including the extra residual down sample layer when using NCSN++) to be tuned.
\begin{table}[b]
\vspace{-.2em}
\centering
\begin{center}
\tablestyle{5pt}{1.05}
\begin{tabular}{x{75}x{78}x{88}x{75}|x{28}}
Norm Layers (7.9\%)& Conv Layers (85.8\%) & QKV Projections (2.1\%) & Skip Layers (4.0\%)&   FID (\(\downarrow\)) \\
\shline
\ding{51} & \ding{51} & \ding{51} & \ding{51}& 1.66 \\
\ding{51} & \ding{55} & \ding{51} &\ding{51}& \textbf{1.54} \\
\ding{51} & \ding{55} & \ding{55} & \ding{51}& 1.60   \\
\ding{51} & \ding{55}  & \ding{55} & \ding{55}& 2.51   \\
\end{tabular}
\end{center}
\hspace{2em}

\vspace{-.1em}
\caption{\textbf{GDD-I: ablation experiments}. \ding{51} indicates the corresponding layer can be tuned and \ding{55} indicates frozen layers. The percentage in the brackets is the proportion of the total parameters in these layers.}
\label{tab:ablations_fz} \vspace{-.5em}
\end{table}

The results confirm our idea (Table \ref{tab:ablations_fz}). Freezing most of the convolutional layers in the UNet blocks not just works. It actually produces better performance. Further freezing the QKV projections (\cite{vaswani2023attention}) causes a slight decrease in the performance, and freezing the skip layers on the residual connections leads to worse results, suggesting that tuning of these two types of layers is necessary for distillation. These results strongly support our theory that the ability for one-step generation is inherent in DMs: with most of the original parameters locked (85.8\% of the parameters contained in the convolutional layers) in the pre-trained score model, our model achieves results better than the previous SOTA results. We term the method that freezes most of the convolutional layers as GDD-I (GAN Distillation at the Distribution level using Innate ability). One of the critical challenges in training GANs is the inherent instability of the training process. However, our method circumvents this issue by freezing the majority of parameters in both the Discriminator and the Generator. As a result, the training process of GDD-I remains remarkably stable, with minimal instances of mode collapse.

\subsection{Extra Instance Level loss}\label{sec:extraCD}
\begin{figure}[hb]
\centering
   \includegraphics[width=\linewidth]{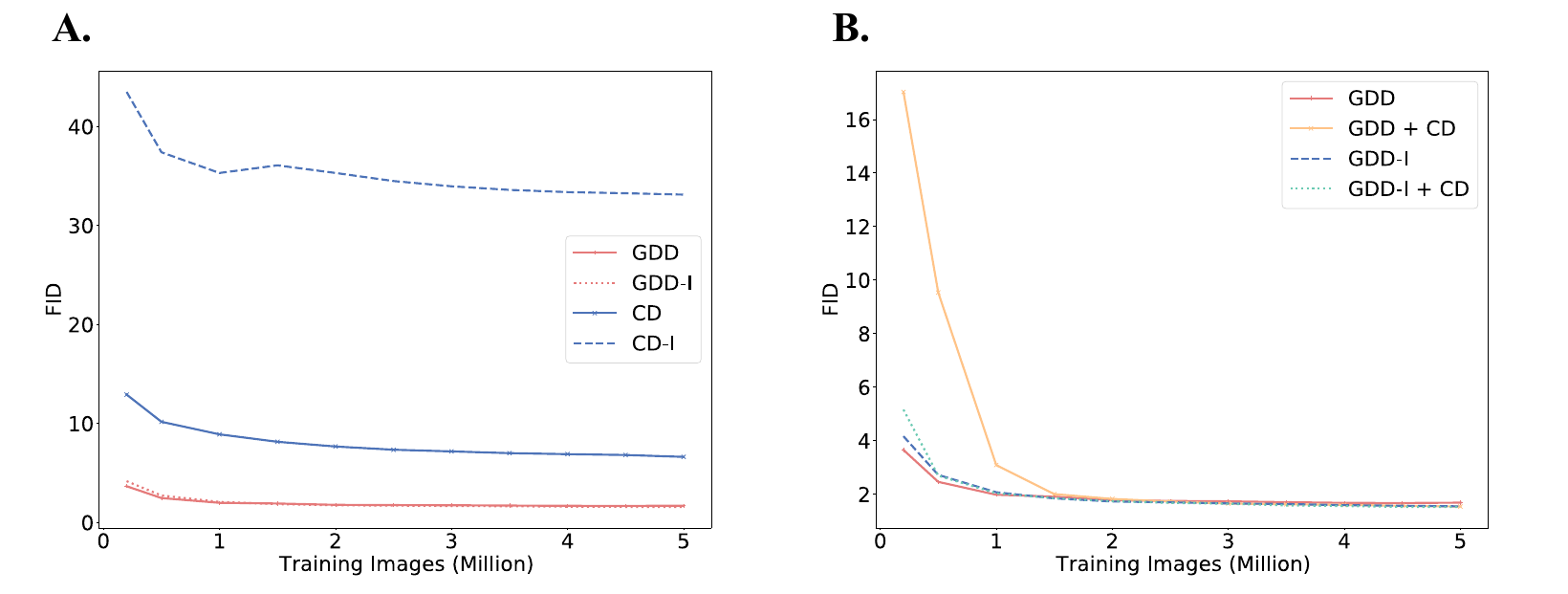}
\caption{Comparison and effect of CD loss. \textbf{A.} Comparison of CD and CD-I. \textbf{B.} Effect of adding an extra CD loss.}
\label{fig:CDLoss}
\end{figure}

In addition, we test the model's performance when extra instance-level loss is used. We use CD loss as an example since it is easy to implement and performs well. The comparisons are conducted on CIFAR-10 dataset. We notice that adding an extra CD loss significantly slows down the converging speed in the beginning. It reaches slightly better results when compared with GDD. This advantage disappears when compared against GDD-I (Figure \ref{fig:CDLoss} \textbf{B.}). Note that adding an extra CD loss almost doubles the training resource and time.

We further compare our methods with the original CD and CD with frozen convolutional layers (CD-I) (Figure \ref{fig:CDLoss} \textbf{A.}). CD-I results in poorer performance, suggesting that forcing the student model to estimate the teacher model requires adjustments in the convolutional layers. This further proves our findings in Section \ref{sec:instance}. It is still possible to combine our method with other instance-based distillation methods (e.g. PD, CTM, SiD (\cite{zhou2024score})). We leave this investigation for future endeavors.

\subsection{Comprehensive Performance}

\paragraph*{Basic Setting.}
Finally, we perform a comprehensive test of both GDD and GDD-I on CIFAR-10, AFHQv2 64x64  (\cite{choi2020stargan}), FFHQ 64x64 (\cite{karras2019stylebased}), and ImageNet 64x64 (\cite{5206848}). The results for class-conditional generation are reported on CIFAR-10 and ImageNet 64x64. The pre-trained diffusion models are from EDM. We use the ADM (\cite{dhariwal2021diffusion}) architecture for ImageNet and the NCSN++ architecture for the other datasets. We use lr=1e-4 for CIFAR-10, lr=2e-5 for AFHQv2 64x64 and FFHQ 64x64, and lr=8e-6 for ImageNet. Batch size is set to 512 for ImageNet and 256 for all other datasets. More details can be found in Appendix \ref{sec:details}. We report FID for all datasets, Inception Score (IS, \cite{salimans2016improved}) for CIFAR-10, precision and recall (\cite{kynkäänniemi2019improved}) for ImageNet 64x64. For the discriminator, we use VGG16 with batch normalization and EfficientNet-lite0 as the feature network for CIFAR-10, DeiT (\cite{touvron2021training}) and EfficientNet-lite0 for the other datasets. The optimal settings in Section \ref{sec:EXP1} is used.

\begin{table*}[t]\tiny
\centering
\subfloat[\textbf{CIFAR-10}]{
\begin{minipage}{1.0\linewidth}
\begin{center}
\tablestyle{3pt}{1}
\begin{tabular}{y{145}x{30}x{5}x{28}x{28}x{5}x{35}}
 \multirow{2}{*}{Method} & \multirow{2}{*}{NFE (\(\downarrow\))} & & \multicolumn{2}{c}{Unconditional} &  & \multicolumn{1}{c}{Conditional} \\
  \cline{4-5} \cline{7-7}
  & & & FID (\(\downarrow\)) & IS (\(\uparrow\)) & & ~~FID (\(\downarrow\)) \\
 \shline
 \vspace{0.5pt}
 \textbf{Direct Generation} & & && & & \\
 \shline
  DDPM (\cite{ho2020denoising})             & 1000 && 3.17 & &&  \\
  DDIM (\cite{song2022denoising}            & 100  && 4.16 & &&   \\
  Score SDE (\cite{song2021scorebased})     & 2000 &&      & && 2.20  \\
  DPM-Solve-3 (\cite{lu2022dpmsolver})      & 48   &&      & && 2.65 \\
  EDM (\cite{karras2022elucidating})        & 35   && 1.98 & && 1.79 \\
  BigGAN (\cite{brock2019large})            & 1    &&      & && 14.73 \\
 StyleGAN2-ADA (\cite{karras2020training}) & 1     && 2.92 & 9.82 && 2.42   \\
 \vspace{0.5pt}
 \textbf{Distillation} &&&&&& \\
 \shline
 PD (\cite{salimans2022progressive})        & 1    && 9.12 &  &&  \\
 DFNO (\cite{zheng2023fast})                & 1    && 3.78 &  && \\
 CD (\cite{song2023consistency})            &1     && 3.55 &  &&  \\
 CD (\cite{song2023consistency})            &2     && 2.93 &  &&  \\
 iCT (\cite{Song2023ImprovedTF})            &1     && 2.83 & 9.54 && \\
 iCT (\cite{Song2023ImprovedTF})            &2     && 2.46 & 9.80 && \\
 iCT-deep (\cite{Song2023ImprovedTF})       &1     && 2.51 & 9.76 && \\
 iCT-deep (\cite{Song2023ImprovedTF})       &2     && 2.24 & 9.89 && \\
 CTM (\cite{kim2024consistency})            & 1    && 1.98 &      && 1.73   \\
 CTM (\cite{kim2024consistency})            & 2    && 1.87 &      && 1.63 \\
 DMD (\cite{yin2023onestep})                & 1    && 2.62 &      && \\
 SiD (\(\alpha=1\), \cite{zhou2024score})   & 1    && 2.02 & 10.01 && 1.93  \\
 SiD (\(\alpha=1.2\), \cite{zhou2024score}) & 1    && 1.92 & 9.98  && 1.71   \\
 GDD (ours)                                 & 1    && 1.66 &\textbf{10.11} && 1.58\\
 GDD-I (ours)                               & 1   && \textbf{1.54} & 10.10 && \textbf{1.44}  \\
\end{tabular}
\end{center}
\end{minipage}}
\vspace{15pt}
\parbox[b]{\textwidth}{%
\subfloat[][\textbf{AFHQv2 64x64}, unconditional]{
\begin{minipage}{0.5\linewidth}
\begin{center}
\tablestyle{3pt}{1}
\begin{tabular}{y{120}x{30}x{28}}
  Method & NFE (\(\downarrow\)) & FID (\(\downarrow\))  \\
 \shline
\vspace{0.5pt}
 \textbf{Direct Generation} & &   \\
 \shline
 EDM (\cite{karras2022elucidating}) & 79 & 1.96  \\
 \vspace{0.5pt}
 \textbf{Distillation} & &    \\
 \shline
SiD (\(\alpha=1.0\), \cite{zhou2024score}) & 1 & 1.62 \\
SiD (\(\alpha=1.2\), \cite{zhou2024score}) & 1 & 1.71 \\
GDD (ours) &  1 &  \textbf{1.23} \\
GDD-I (ours) &  1 & 1.31 \\
\end{tabular}
\end{center}
\end{minipage}}
\subfloat[][\textbf{FFHQ 64x64}, unconditional]{
\begin{minipage}{0.5\linewidth}
\begin{center}
\tablestyle{3pt}{1}
\begin{tabular}{y{120}x{24}x{24}}
  Method & NFE(\(\downarrow\)) & FID(\(\downarrow\))  \\
   \shline
  \vspace{0.5pt}
 \textbf{Direct Generation} & &   \\
 \shline
  EDM  (\cite{karras2022elucidating})& 79 & 2.39 \\
   \vspace{0.5pt}
 \textbf{Distillation} & &   \\
 \shline
 BOOT (\cite{gu2023boot}) & 1 & 9.0 \\
 SiD ($\alpha=1.0$, \cite{zhou2024score})& 1 & 1.71 \\
SiD ($\alpha=1.2$, \cite{zhou2024score})& 1 & 1.55 \\

GDD(ours) &  1& 1.08  \\
 GDD-I(ours) &  1& \textbf{0.85} \\
\end{tabular}
\end{center}
\end{minipage}   
}}
\vspace{10pt}
\vspace{-.1em}
\caption{Comprehensive comparison between GDD, GDD-I, and previoius work. Tests are done on CIFAR-10, AFHQv2 64x64 and FFHQ 64x64.}
\label{tab:results} \vspace{-.5em}
\end{table*}

\begin{table*}[t]
\vspace{-.2em}
\centering
\begin{minipage}{1\linewidth}
\begin{center}
\tablestyle{3pt}{1}
\begin{tabular}{y{135}x{30}x{30}x{32}x{30}}
     Method & NFE (\(\downarrow\)) & FID (\(\downarrow\)) & Prec. (\(\uparrow\)) & Rec. (\(\uparrow\)) \\
      \shline
      \vspace{0.5pt}
 \textbf{Direct Generation} & &   \\
 \shline
     RIN (\cite{jabri2023scalable}) &1000 & 1.23 &  \\
     DDPM (\cite{ho2020denoising})  &250 & 11.00 & 0.67 & 0.58  \\
     ADM (\cite{dhariwal2021diffusion}) & 250& 2.07 & 0.74 & 0.63 \\
     EDM (\cite{karras2022elucidating}) & 79 & 2.64 &      &  \\
     iCT (\cite{Song2023ImprovedTF})  & 1 & 4.02 & 0.70 & 0.63 \\
     iCT (\cite{Song2023ImprovedTF})  & 2 & 3.20 & 0.73 & 0.63 \\
     iCT-deep (\cite{Song2023ImprovedTF}) & 1& 3.25 & 0.72 & 0.63 \\
     iCT-deep (\cite{Song2023ImprovedTF}) & 2& 2.77 & 0.74 & 0.62 \\
    BigGAN-deep (\cite{brock2019large}) & 1& 4.06& 0.79 & 0.48  \\
     StyleGAN2-XL (\cite{sauer2022styleganxl}) &1  & 1.51 &  \\

\vspace{0.5pt}
 \textbf{Distillation} & &   \\
 \shline
     PD (\cite{salimans2022progressive}) &1 & 15.39 &  \\
     BOOT (\cite{gu2023boot}) &1 & 16.3 & 0.68& 0.36 \\
     DFNO (\cite{zheng2023fast}) & 1& 7.83 & & 0.61 \\
     CD (\cite{song2023consistency}) & 1& 6.20 &  0.68 & 0.63 \\
     CD (\cite{song2023consistency}) & 2& 4.70 & 0.69  & 0.64 \\
     CTM (\cite{kim2024consistency})& 1& 1.92  & 70.38 & 0.57 \\
     CTM (\cite{kim2024consistency})& 2& 1.73 & 64.29 & 0.57 \\
     DMD (\cite{yin2023onestep}) & 1 & 2.62 & & \\
     SiD ($\alpha$=1, \cite{zhou2024score}) & 1 & 2.02 & 0.73 &0.63 \\
    SiD ($\alpha$=1.2, \cite{zhou2024score}) & 1 & 1.52 & 0.74 &0.63 \\
     GDD(ours) & 1 & 1.42 & \textbf{0.77} & 0.59  \\
      GDD-I(ours) & 1 & \textbf{1.16} &  0.75 & 0.60  \\
\end{tabular}
\end{center}
\end{minipage}
\caption{Comprehensive comparison between GDD, GDD-I, and previous work. Tests are done on ImageNet 64x64, class-conditional.}
\label{tab:results_imgnet} \vspace{-.5em}
\end{table*}

\paragraph*{Results.}
The results can be found in Table \ref{tab:results} and \ref{tab:results_imgnet}. Both GDD and GDD-I achieve the SOTA results on all datasets that we have tested. Our methods are not only suitable for unconditional distillation, but also show great performance on class-conditional generation tasks. Moreover, GDD-I outperforms GDD by a significant margin in most of the datasets, except AFHQv2 64x64. On CIFAR-10, our methods achieve the best FID of 1.54 and the best IS of 10.11. Similarly, superior performance is found on AFHQv2 64x64 (FID=1.23) and FFHQ 64x64 (FID=0.85). In ImageNet 64x64, the model also achieves SOTA performance (FID=1.16, precision=0.77). Together, these results demonstrate the robustness and effectiveness of our methods. They support our hypothesis that DMs are capable of generating results in one-step innately.

\section{Limitations}
While our approach shows great performance and speed, our experiments are mainly done on datasets with relatively low resolution. Performance on high resolution needs to be further verified. In Section \ref{sec:Innate}, our experiments show that DMs are differentially activated at different time steps in UNet (\cite{ronneberger2015unet}) style architectures. However, the specific roles of these layers in multi-step generation and one-step generation tasks remain to be explored. Additionally, further experiments are needed to determine whether this occurs in other architectures as well (e.g. DiT (\cite{peebles2023scalable}). Finally, while our method achieves high sample quality for both unconditional and class-conditional generation, functions like image-editing and inpainting are yet to be implemented. A method that integrates the quality and efficiency of our approach and the diverse capabilities of instance-based methods remains to be explored.

\section{Conclusion}
In this work, we have identified an inherent limitation in instance-based distillation methods, stemming from the disparities between the teacher and the student models. To circumvent this limitation, we first propose a novel method, GDD (GAN Distillation in Distribution level), which leverages an exclusive distributional loss to efficiently distill a one-step generator from diffusion models. We further investigate the activation pattern of diffusion model layers and develop GDD-I (GDD using Innate ability), which enhances performance by freezing most of the convolutional layers in the GDD model during the distillation process and outperforms the original GDD method. Our work not only establishes new SOTA results with minimal computational resources but also elucidates a critical principle in the field of diffusion distillation, paving the way for more efficient and effective model training methodologies in image generation and beyond.

\bibliography{gdd}
\bibliographystyle{gdd}

\clearpage

\appendix

\section{Related Works}

\paragraph*{Image Generation.} 
Image generation is an important task that has been explored over the years. GAN(\cite{goodfellow2014generative},\cite{brock2019large},\cite{karras2020analyzing}) models have dominated the field for a long time. Quantization-Based Generative Model(\cite{esser2021taming}, \cite{chang2022maskgit},\cite{li2023mage}) first encodes images to discrete tokens, then use a transformer to model the probability distribution between those tokens. Diffusion Models(\cite{sohldickstein2015deep}, \cite{song2020generative}, \cite{ho2020denoising}, \cite{song2020improved}, \cite{song2021scorebased}) or score-based generative models, tries to learn an accuracy estimation of scores (the gradient of the log probability density), from a perturbed distribution with a Gaussian kernel to the image distribution. Diffusion Models achieve great success especially in image generation(\cite{dhariwal2021diffusion},\cite{nichol2022glide},
\cite{ramesh2022hierarchical},\cite{saharia2022photorealistic}) and are widely used in different fields.

\paragraph*{Diffusion Distillation.} One of the challenges of diffusion models in practice is the high computational cost incurred during fine multi-step generation. A series of distillation methods have been proposed to distill diffusion models to one-step generators. Some distillation methods (\cite{salimans2022progressive},\cite{song2023consistency} keep the original features of the diffusion models, such as image inpainting and image editing. Others (\cite{zhou2024score}) focus on improving the one-step generation results. Distillation models with GAN were introduced recently. GAN loss has been used as an auxiliary loss of instance-level distillation loss (\cite{kim2024consistency}), or a replacement of instance-level loss (L2 or LPIPS) as in \cite{lin2024sdxllightning}. Both of these works achieve great performance, showing the potential of GAN-Based distillation. 

\section{Appendix}
\subsection{Definition of Instance-Based Distillation}\label{sec:instance_details}
\begin{table*}[hbt]
\vspace{-.2em}
\centering
\subfloat[\textbf{Sampling strategies for time steps}.]{
\begin{minipage}{\linewidth}
\begin{center}
\tablestyle{6pt}{1.25}
    \begin{tabular}{y{25}|x{75}x{75}x{92}}
    Method & \(\mathbf{t}\) & \(\mathbf{u}\) & \(\mathbf{s}\)   \\ 
     \shline
     PD  & \(T(i,N), 2 \leq i \leq N\) & \(T(\frac{1}{2}(i+j),N)\) & \(T(j,N), 0\leq j<(i-1)\) \\
     CD & \(T(i,N), 1\leq i \leq N\) & \(T(i-1,N)\) & \(T(0,N)\) \\
     CTM & \(T(i,N), 2\leq i \leq N\) & \(T(j,N), 1\leq j \leq i\) & \(T(k,N), 0\leq k \leq j\)  \\
  \end{tabular}
\end{center}
\end{minipage}}
\vspace{10pt}
\subfloat[\textbf{Teacher and student models}.]{
\begin{minipage}{0.88\linewidth}
\begin{center}
\tablestyle{6pt}{1.25}
    \begin{tabular}{y{25}|x{100}x{190}}
    Method & Student Model (\(\mathbf{F}_{\theta}\)) & Teacher Model (\(\mathbf{H}\))   \\ 
     \shline
     PD  & \(\displaystyle \frac{t-s}{t} (\mathbf{f}_{\theta}(x_t,t) - x_t)+x_t\) &  \(\Big\{\begin{array}{lr}
        \mathbf{Solver}_\phi(\mathbf{Solver}_{\phi}(x_t,t,u),u,s), &   N_i=N_{max} \\
       \mathbf{F}^{sg}_\theta(\mathbf{F}^{sg}_{\theta}(x_t,t,u),u,s), &  N_i\neq N_{max}
        \end{array}\)\\
     CD & \(\displaystyle \frac{t-s}{t}(\mathbf{f}_{\theta}(x_t,t) - x_t)+x_t\)&  \(\mathbf{F}^{sg}_\theta(\mathbf{Solver}_{\phi}(x_t,t,u),u,s)\)  \\
     CTM  & \(\displaystyle \mathbf{f}_{\theta}(x_t,t,s)\) & \(\mathbf{F}^{sg}_\theta(\mathbf{Solver}\footnote{CTM use an iterative solver here}_\phi(x_t,t,u),u,s)\) \\
  \end{tabular}
\end{center}
\end{minipage}}
\vspace{10pt}
\subfloat[\textbf{Loss function}.]{
\begin{minipage}{\linewidth}
\begin{center}
\tablestyle{6pt}{1.25}
    \begin{tabular}{y{25}|x{215}}
    Method & Loss Function (\(\mathcal{L}\)) \\ 
     \shline
     PD  &  \(\mathbf{D}(\mathbf{f}_\theta(x_t,t), \frac{t}{t-s}(\mathbf{H}(x_t, t, u, s) - x_t)+x_t)\) \\
     CD  & \(\mathbf{D}(\mathbf{F}_\theta(x_t,t,s), \mathbf{H}(x_t, t, u, s))\)\\
     CTM  &\(\mathbf{D}(\mathbf{F}^{sg}_\theta(\mathbf{F}_\theta(x_t,t,s), s, \sigma_{min}), \mathbf{F}^{sg}_\theta(\mathbf{H}(x_t, t, u, s),s, \sigma_{min}))\)\\
  \end{tabular}
\end{center}
\end{minipage}}
\caption{Definition of commonly used instance-level distillation methods}
\label{tab:dist_method}  
\end{table*}
Progressive Distillation (PD, \cite{salimans2022progressive}) uses a  progressive distillation strategy. Using \(\mathbf{G}_\phi\) with a sufficient number of steps, e.g., \(N_{max}=1024\) as a teacher, PD first distills a student model \(\mathbf{F}_\theta\) with fewer steps, typically \(N_{1}=\frac{1}{2}N_{max}=512\). Then with the fixed student model \(\mathbf{F}^{sg}_\theta\) as teacher where \textit{sg} means stop gradient, PD further distills the student with \(N_{2}=\frac{1}{4}N_{max}=256\). This progress is repeated until a one-step generator is produced.

Consistency Distillation (CD, \cite{song2023consistency}) and Consistency Trajectory Model (CTM, \cite{kim2024consistency}) are similar. Both of these two methods adopt a joint training strategy in which student model \(\mathbf{F}_\theta\) is used to be part of the teacher model \(\mathbf{H}(x_t, t, u, s) = \mathbf{F}^{sg}_{\theta}(\mathbf{G}_\phi(x_t, t, u), u, s)\) at the same time. CTM further projects \(v_s\) and \(\Bar{v}_s\) to the pixel space by \(\mathbf{F}^{sg}_\theta(v_s, s, \sigma_{min})\) and \(\mathbf{F}^{sg}_\theta(\Bar{v}_s, s, \sigma_{min})\) and then calculates \(\mathbf{D}(v_{\sigma_{min}},\Bar{v}_{\sigma_{min}})\) as loss.

The three models are summarized in Table \ref{tab:dist_method}. We modify the definition of PD with a EDM manner here to be consistent. \(t\), \(u\), \(s\) are the start, intermediate and end time steps of the trajectory, respectively. PD sets the \(u\) to the "middle" of \(t\) and \(s\). CD sets the \(u\) as the successor of \(t\) and fixes the \(s\) to zeros. PD uses a separate training strategy since the teacher model can be different with different \(N\)s, and PD supervises \(f_\theta\) with the intersection of \(x=0\) and \(\text{Slope}(x_t,\Bar{v}_s)\), where \(\text{Slope}(x_t,\Bar{v}_s)\) is the slope between \(x_t\) and \(\Bar{v}_s\). CTM first translates \(v_s\) and \(\Bar{v}_s\) to \(v_{\sigma_{min}}\) and \(\Bar{v}_{\sigma_{min}}\), then calculates the loss.

\subsection{Time Step Scheduler}\label{sec:time_step}
When training with CD loss, we use Heun solver and set the total number of time steps to 18 (\(N=18\)) following CD. The time step scheduler is slightly different from the scheduler defined in EDM (\cite{karras2022elucidating}). We reverse the index and define it as a function of current step and total steps:
\begin{equation}
    T(i,N) = \left( {\sigma_{max}}^\frac{1}{\rho} + \Big( 1-\frac{i}{N} \Big) \left( {\sigma_{min}}^\frac{1}{\rho} - {\sigma_{max}}^\frac{1}{\rho} \right) \right)^\rho
\end{equation}
Following Consistency Model, we set \(\rho=7\), \(\sigma_{min}=0.002\) and \(\sigma_{max}=80\).

\subsection{Details for Calculating rel-FID}\label{sec:rel-FID_details}
Both the teacher and the student model are  initialized from a well-trained score model in EDM. To ensure that each model performs well, we adopt the same optimal setting in Section \ref{sec:EXP1} during training. All teacher models reach FID\(\leq\)2.2 in a few training steps. The rel-FIDs are then calculated for these model as described in Section \ref{sec:instance}.

\subsection{Implementation Details}\label{sec:details}

\begin{table}[hbt]
    \caption{Hyperparameters used for training GDD and GDD-I}\label{tab:hyperparameters}
    \centering
        \begin{tabular}{l|c|c|c|c}
            \shline
            Hyperparameter & CIFAR-10 & AFHQv2 $64\times 64$  & FFHQ $64\times 64$ & ImageNet $64\times 64$ \\
            \hline
            G Architecture & NSCN++ &  NSCN++ & NSCN++ & ADM \\
            G LR & 1e-4 &  2e-5 & 2e-5& 8e-6\\
            D LR & 1e-4 & 4e-5 & 4e-5 & 4e-5\\
            Optimizer & Adam & Adam & Adam & Adam \\
            \(\beta_1\) of Optimizer & 0 & 0 & 0 & 0\\
            \(\beta_2\) of Optimizer & 0.99 & 0.99 & 0.99 & 0.99\\
            Weight decay & No & No & No & No\\
            Batch size & 256 & 256 & 256  & 512\\
            $\gamma_{r1}$ (Regularization) & 1e-4 & 1e-4  & 1e-4 & 4e-4 \\
            EMA half-life (Mimg) & 0.5 & 0.5 & 0.5  & 50 \\
            EMA warmup ratio& 0.05 &  0.05 &  0.05 & 0.05 \\
            \multirow{2}{*}{Training Images (Mimg)} & 5 (GDD) & 5 (GDD) & 5 (GDD) & 5 (GDD) \\
            & 5 (GDD-I) & 10 (GDD-I) & 10 (GDD-I) & 5 (GDD-I) \\
            Mixed-precision (BF16) & Yes & Yes & Yes & Yes \\
            Dropout probability & 0.0 & 0.0 & 0.0 & 0.0\\
            Augmentations & No & No & No & No\\
            Number of GPUs & 8 & 8 & 8 & 8 \\
            \shline
        \end{tabular}
\end{table}
\paragraph{Parameterization.}
We follow the parameterization in Consistency Model (\cite{song2023consistency}) for all of our models:
\begin{align*}
    c_\text{skip}(t) = \frac{\sigma_\text{data}^2}{(t-\epsilon)^2 + \sigma_\text{data}^2},\quad c_\text{out}(t) = \frac{\sigma_\text{data} (t-\epsilon) }{\sqrt{\sigma_\text{data}^2 + t^2}}
\end{align*}

We set \(t=\sigma_{max}\) when applying distributional distillation with \(c_\text{skip}\approx 0\) and \(c_\text{out}\approx1\).

\paragraph{Training Details.}

We distill EDMs for all datasets and methods for fair comparison. EMA decay is applied to the weights of generator for sampling following previous work. EMA decay rate is calculated by EMA Halflife and EMA warmup is used as in EDM. EMA warmup ratio is set to 0.05 for all datasets, and this leads to a gradually grow EMA decay rate. We use Adam optimizer with \(\beta_1=0, \beta_2=0.99\) without weight decay for both the generator and the discriminator. No gradient clip is applied. For class-conditional generation, we use no classifier-guidance, but using simple class embedding in discriminator following StyleGAN2-XL. Mixed precision training is adopted for all experiments with \textit{BFloat16} data type, and results with \textit{Float16} and \textit{Float32} are similar. We apply no learning rate scheduler. Images are resized to 224 first before they are fed to the PG discriminator or CD-LPIPS loss. All models are trained on a cluster of Nvidia A100 GPUs. Hyperparameters used for GDD and GDD-I training can be found in Table \ref{tab:hyperparameters}. GDD and GDD-I share the same hyperparameters except for the number of training images, since GDD-I will not overfit when GDD gets worse FID.

\clearpage
\section{Samples}
We include EDM's results for comparison. The initial noises are same as different models. It is clear that GDD and GDD-I generate images similar to the original model (EDM), but they are not exactly the same since our methods allow the student model to perform differently with the teacher model.

\begin{figure}[ht]
\centering
    \includegraphics[width=\linewidth]{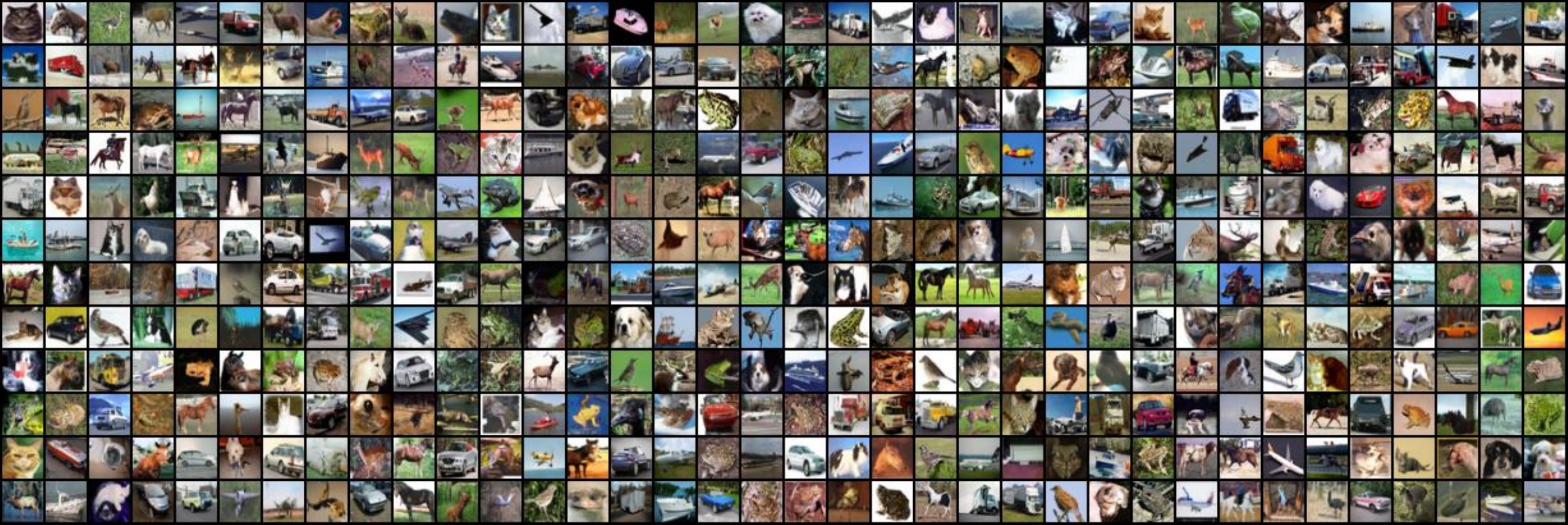}
   \caption{CIFAR-10, EDM, NFE=18, FID=1.96}
\end{figure}

\begin{figure}[ht]
\centering
    \includegraphics[width=\linewidth]{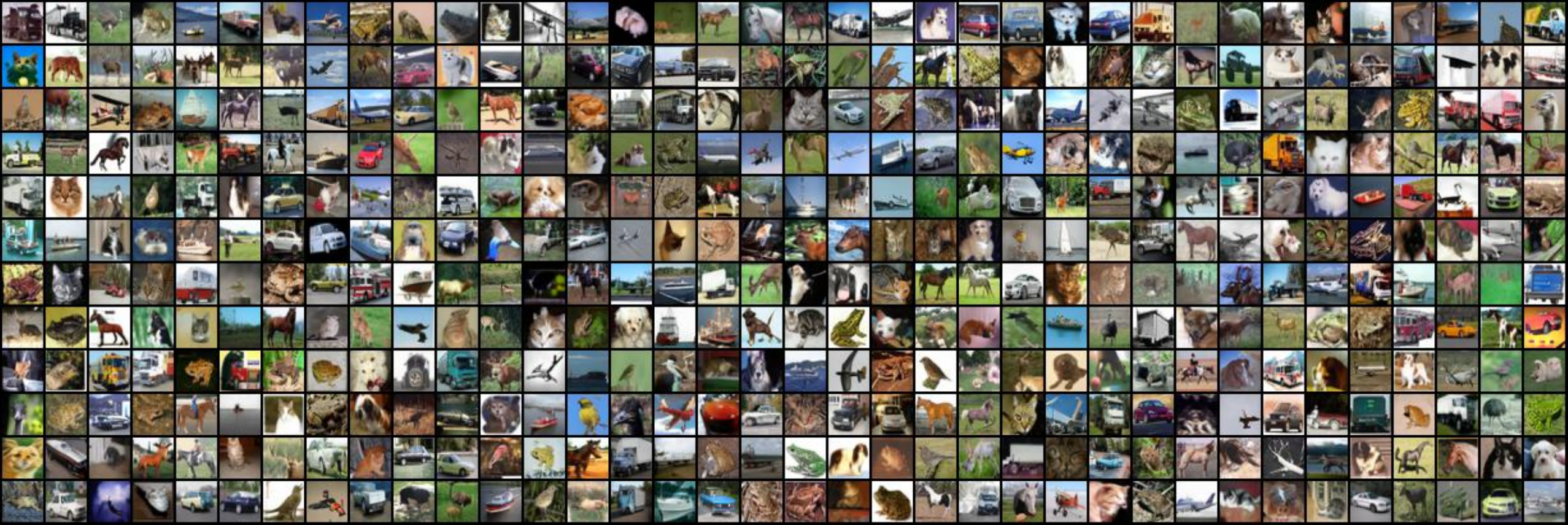}
   \caption{CIFAR-10, GDD, NFE=1, FID=1.66}
\end{figure}

\begin{figure}[ht]
\centering
    \includegraphics[width=\linewidth]{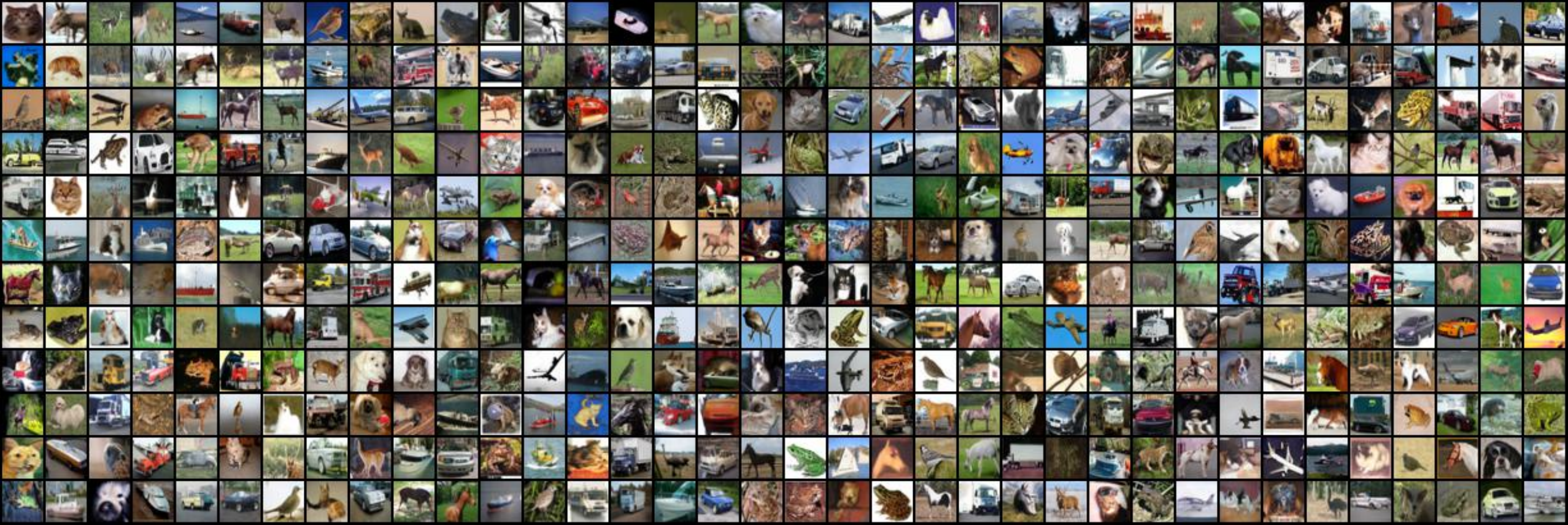}
   \caption{CIFAR-10, GDD-I, NFE=1, FID=1.54}
\end{figure}

\begin{figure}[ht]
\centering
    \includegraphics[width=\linewidth]{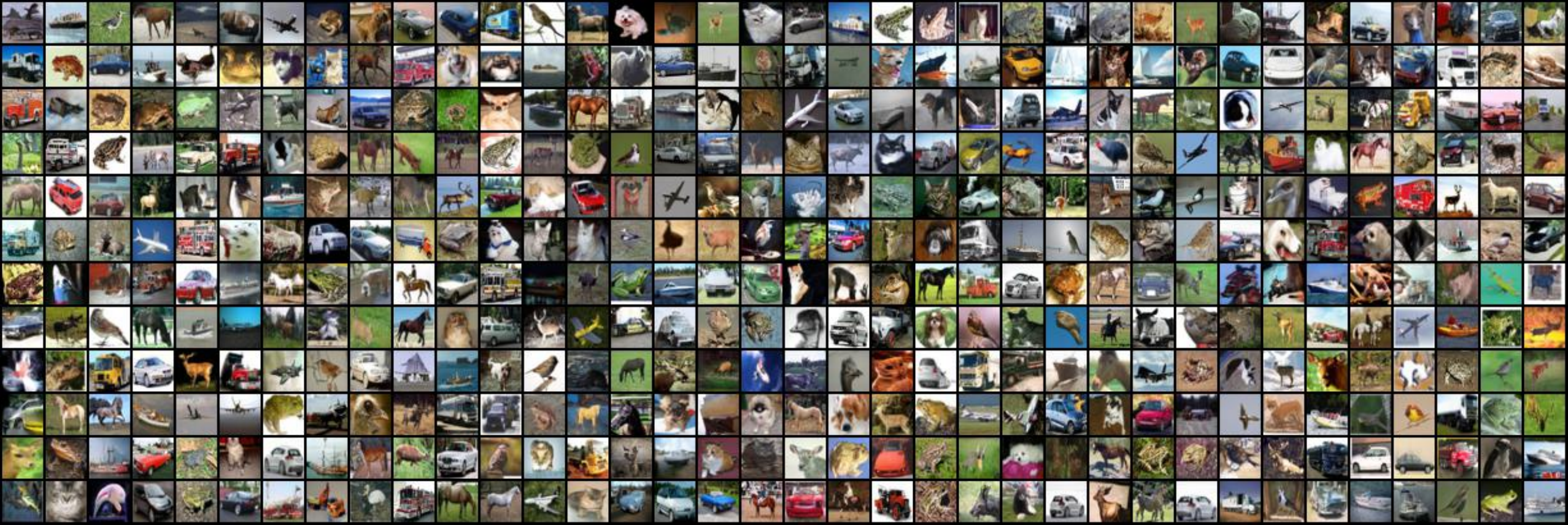}
   \caption{CIFAR-10 (conditional), EDM (VE), NFE=18, FID=1.82}
\end{figure}

\begin{figure}[ht]
\centering
    \includegraphics[width=\linewidth]{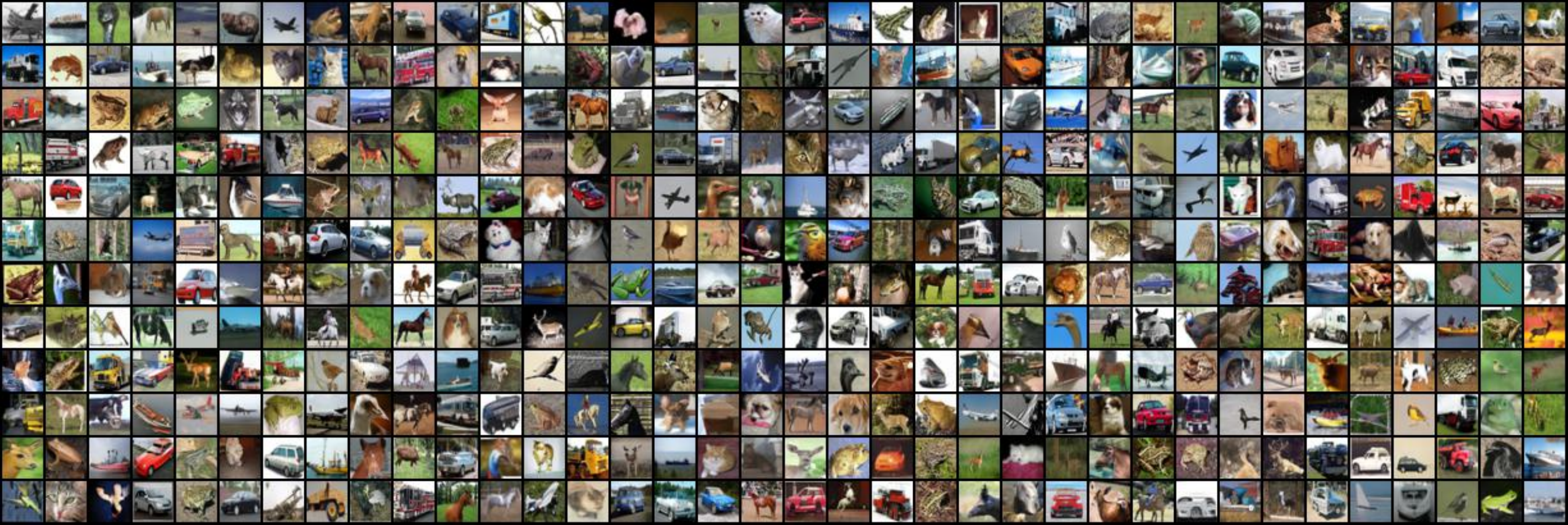}
   \caption{CIFAR-10 (conditional), GDD, NFE=1, FID=1.58}
\end{figure}

\begin{figure}[ht]
\centering
    \includegraphics[width=\linewidth]{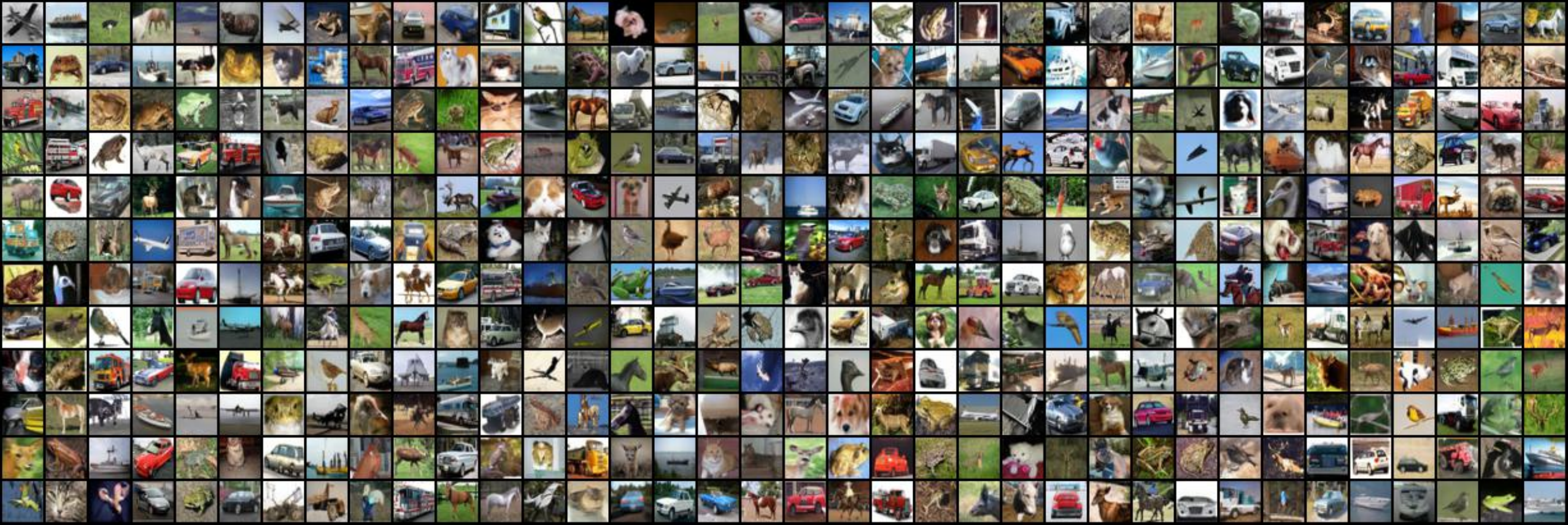}
   \caption{CIFAR-10 (conditional), GDD-I, NFE=1, FID=1.44}
\end{figure}

\begin{figure}[ht]
\centering
    \includegraphics[width=\linewidth]{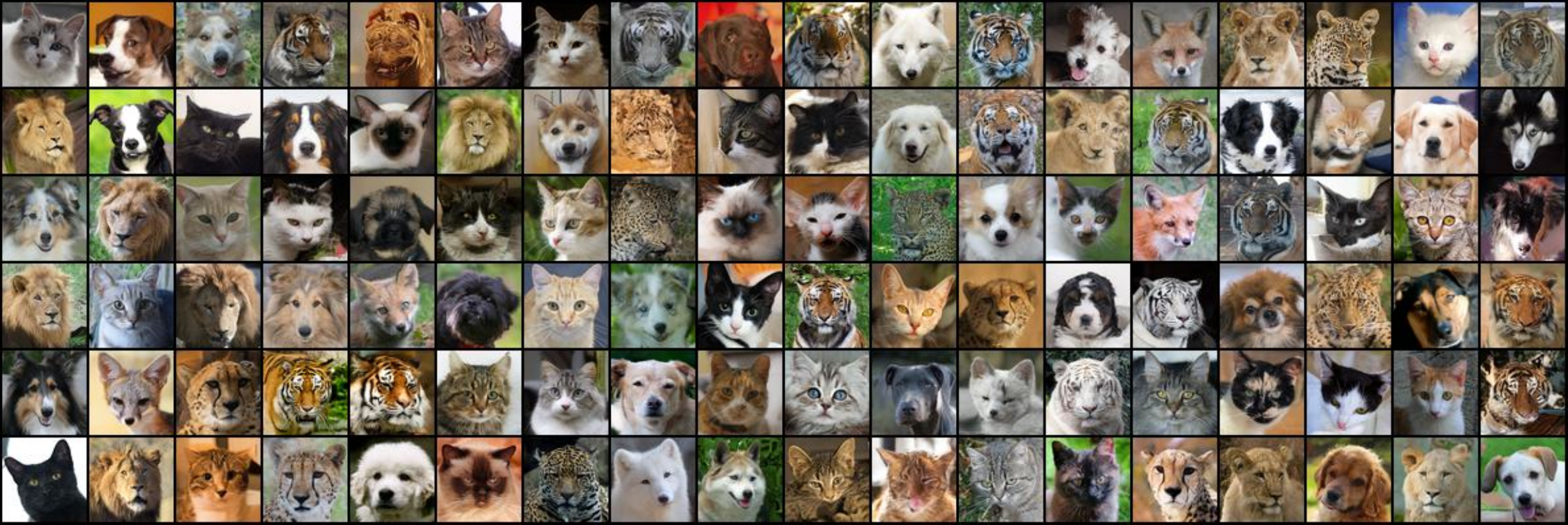}
   \caption{AFHQv2 64x64, GDD, NFE=79, FID=2.17}
\end{figure}

\begin{figure}[ht]
\centering
    \includegraphics[width=\linewidth]{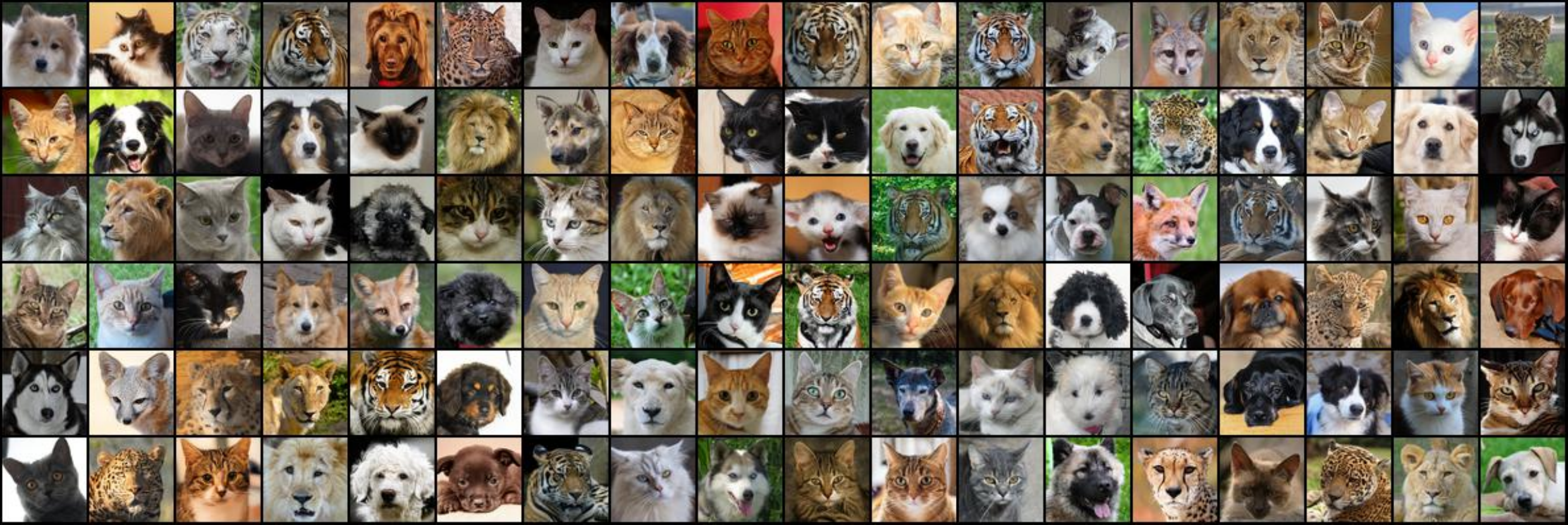}
   \caption{AFHQv2 64x64, GDD, NFE=1, FID=1.23}
\end{figure}

\begin{figure}[ht]
\centering
    \includegraphics[width=\linewidth]{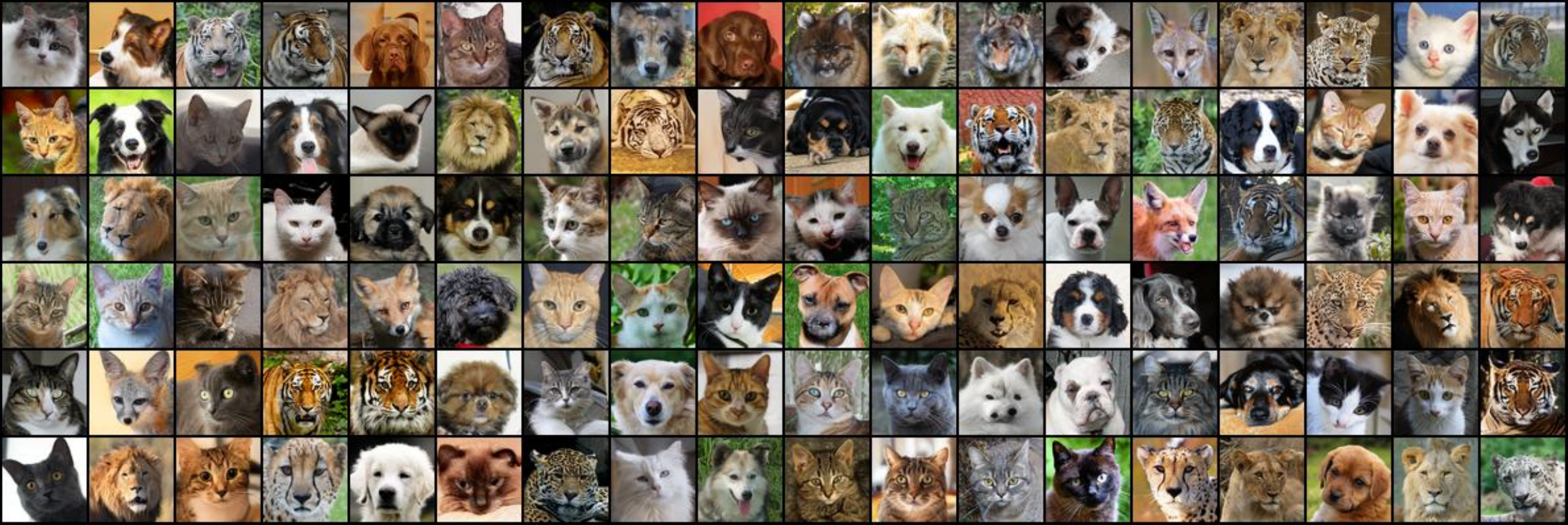}
   \caption{AFHQv2 64x64, GDD-I, NFE=1, FID=1.31}
\end{figure}

\begin{figure}[ht]
\centering
    \includegraphics[width=\linewidth]{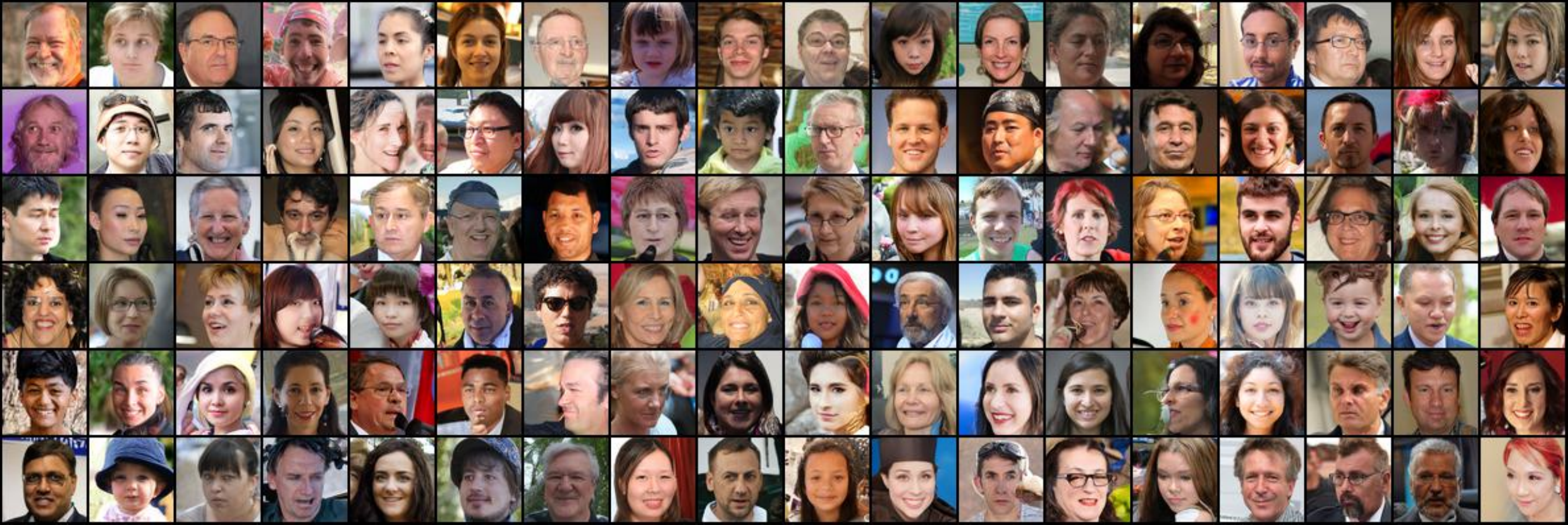}
   \caption{FFHQ 64x64, EDM (VE), NFE=79, FID=2.60}
\end{figure}

\begin{figure}[ht]
\centering
    \includegraphics[width=\linewidth]{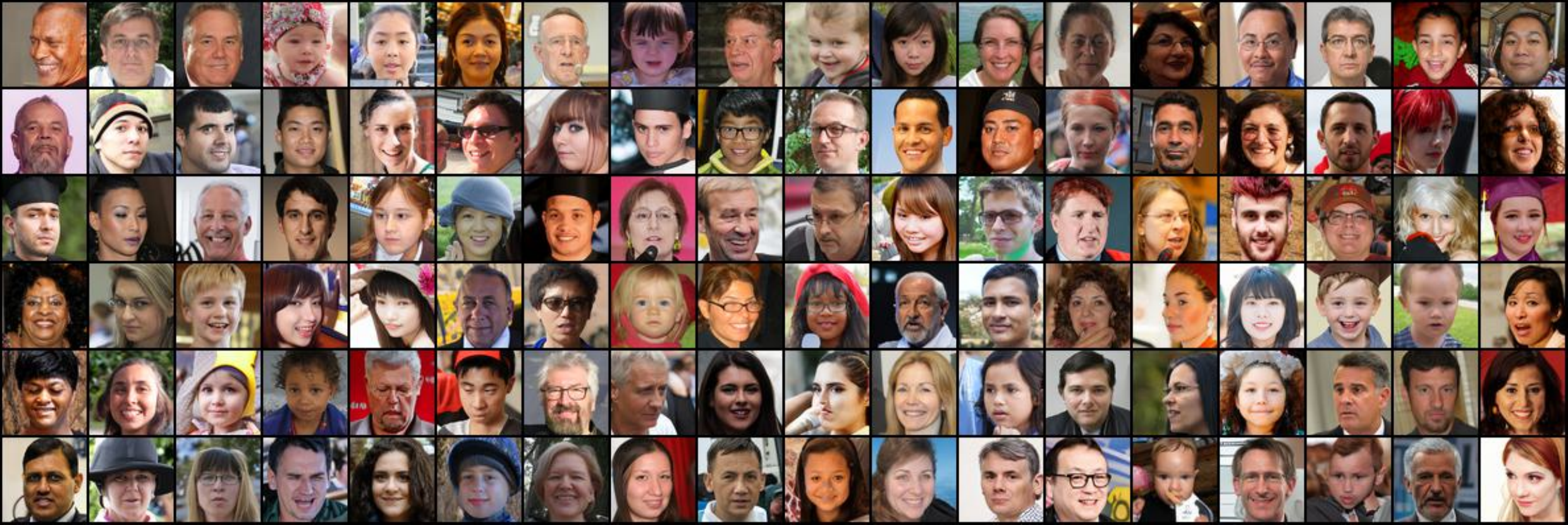}
   \caption{FFHQ 64x64, GDD, NFE=1, FID=1.08}
\end{figure}

\begin{figure}[ht]
\centering
    \includegraphics[width=\linewidth]{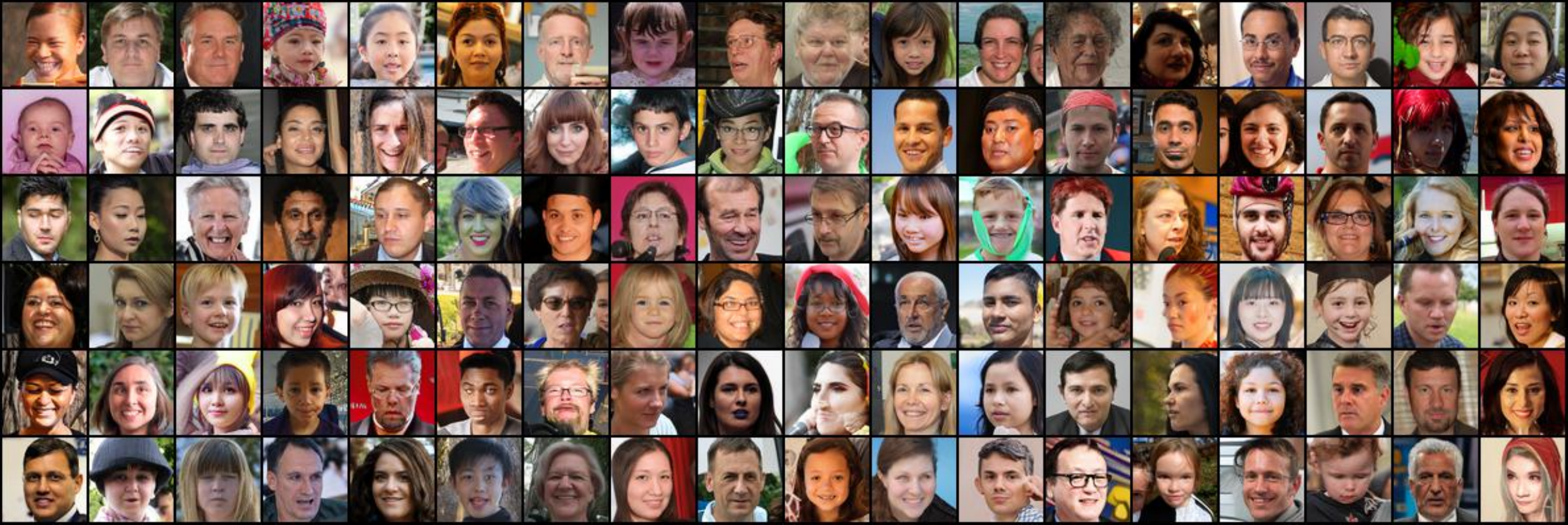}
   \caption{FFHQ 64x64, GDD-I, NFE=1, FID=0.85}
\end{figure}

\begin{figure}[ht]
\centering
    \includegraphics[width=\linewidth]{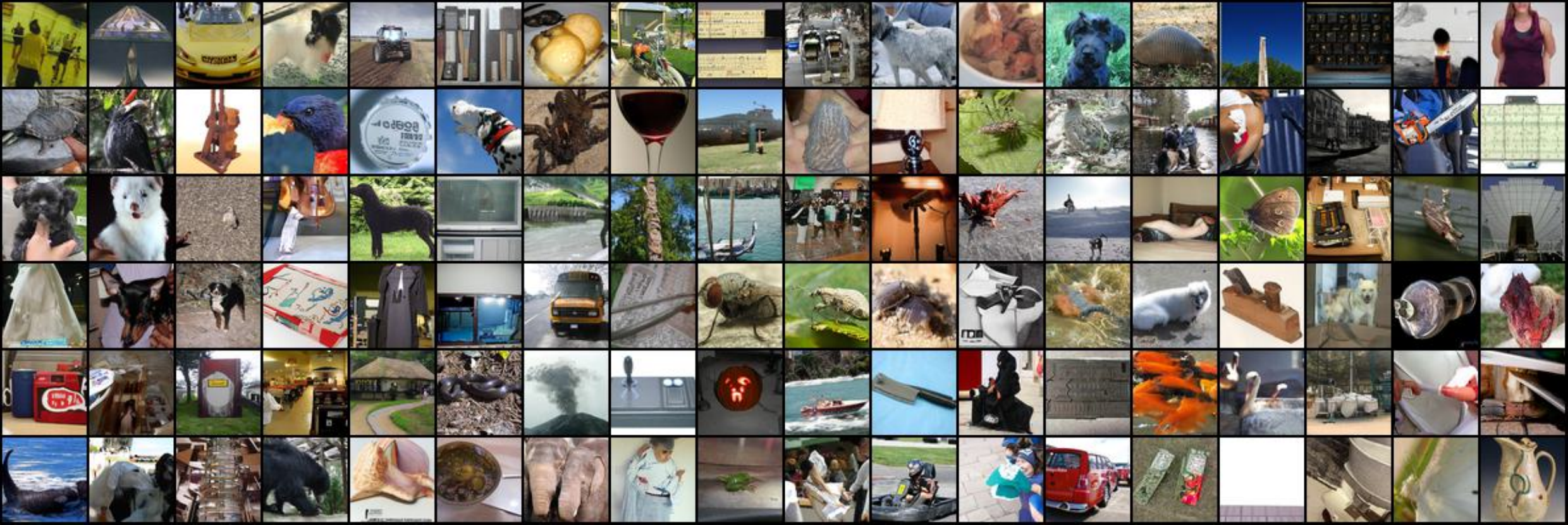}
   \caption{ImageNet 64x64 (conditional), EDM (VE), NFE=79, FID=2.36}
\end{figure}

\begin{figure}[ht]
\centering
    \includegraphics[width=\linewidth]{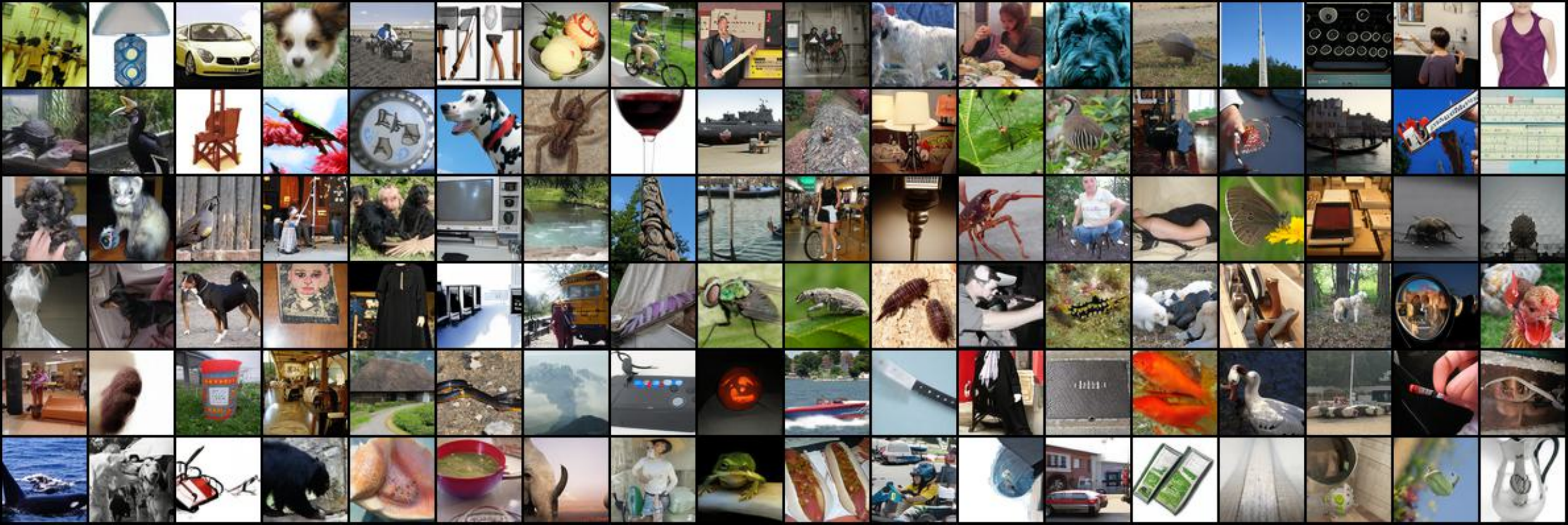}
   \caption{ImageNet 64x64 (conditional), GDD, NFE=1, FID=1.42}
\end{figure}

\begin{figure}[ht]
\centering
    \includegraphics[width=\linewidth]{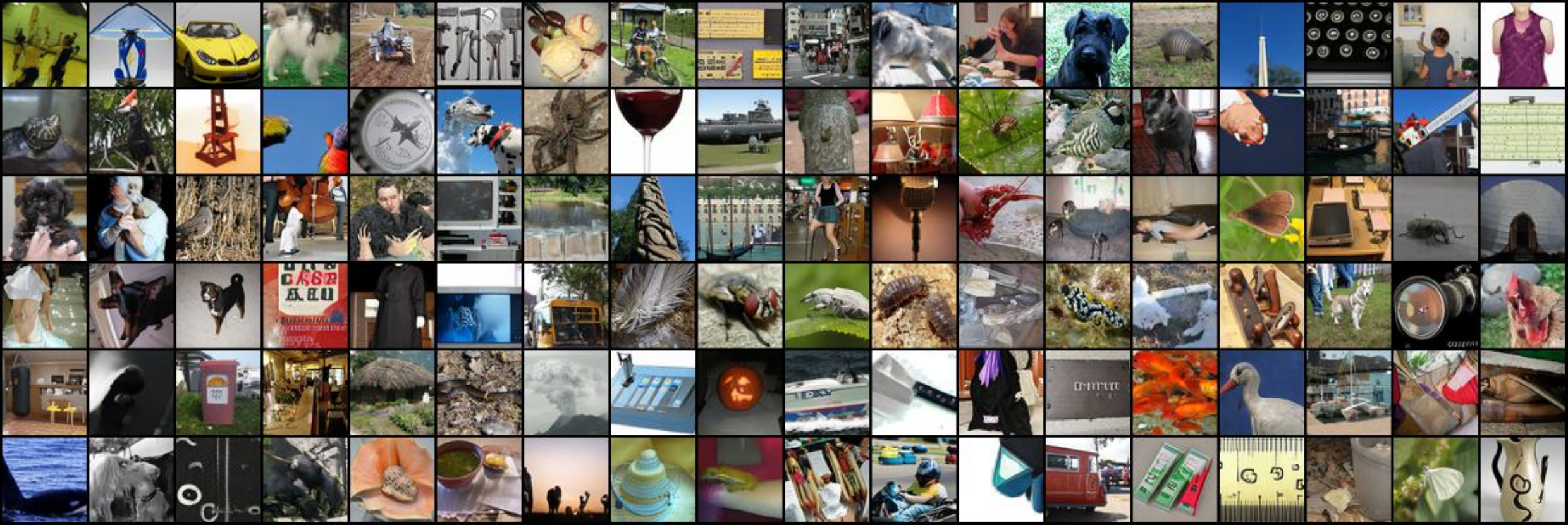}
   \caption{ImageNet 64x64 (conditional), GDD-I, NFE=1, FID=1.16}
\end{figure}

\end{document}